\documentclass[lettersize,journal]{IEEEtran}

\usepackage{amsmath,amssymb,amsfonts}
\usepackage[table]{xcolor}
\usepackage{textcomp}
\usepackage{tikz}
\usepackage{array}
\usepackage{booktabs}
\usepackage{multirow}
\usepackage{graphicx}
\usepackage[most]{tcolorbox}
\usepackage{algorithm}
\usepackage{algorithmic}
\usepackage{enumitem}
\setlist[itemize,2]{label=$\triangleright$}
\usepackage{hyperref}
\usepackage[caption=false,font=normalsize,labelfont=sf,textfont=sf]{subfig}
\usepackage{stfloats}
\usepackage{url}
\usepackage{verbatim}
\usepackage{cite}
\begin{document}

\title{FusionLog: Cross-System Log-based Anomaly Detection via Fusion of General and Proprietary Knowledge}

\author{Xinlong Zhao, Tong Jia, Minghua He, Xixuan Yang, Ying Li
\thanks{Xinlong Zhao, Minghua He, Xixuan Yang and Ying Li are with the School of Software \& Microelectronics, Peking University, Beijing, China (e-mail: xlzhao25@stu.pku.edu.cn; hemh2120@stu.pku.edu.cn; yangxixuan@stu.pku.edu.cn; li.ying@pku.edu.cn).}
\thanks{Tong Jia is with the Institute for Artificial Intelligence, Peking University, Beijing, China (e-mail: jia.tong@pku.edu.cn).}
\thanks{\textit{Corresponding author: Tong Jia.}}
}

\markboth{IEEE Transactions on Services Computing,~Vol.~XX, No.~X, 2026}%
{Zhao \MakeLowercase{\textit{et al.}}: FusionLog: Cross-System Log-based Anomaly Detection}


\maketitle

\begin{abstract}
Log‑based anomaly detection is critical for ensuring the stability and reliability of software systems. One of the key problems in this task is the lack of sufficient labeled logs, which limits the rapid deployment in new systems. Existing works usually leverage large‑scale labeled logs from a mature software system and a small amount of labeled logs from a new system, using transfer learning to extract and generalize general knowledge across both systems. However, these methods focus solely on the transfer of general knowledge and neglect the disparity and potential mismatch between general knowledge and proprietary knowledge of the target system, thus constraining performance. To address this limitation, we propose FusionLog, a novel cross‑system log‑based anomaly detection method that effectively achieves the fusion of cross-system general knowledge and target system's proprietary knowledge. Specifically, we first design a training-free router based on semantic similarity that dynamically partitions unlabeled target logs into “general logs” and “proprietary logs.” For general logs, FusionLog employs a small model based on system‑agnostic representation meta‑learning for direct training and inference, inheriting the general anomaly patterns. For proprietary logs, we iteratively generate pseudo‑labels and fine‑tune the small model using multi-round collaborative knowledge distillation and fusion based on LLM and small model to enhance its capability to recognize target-specific anomaly patterns. We evaluated the performance of FusionLog on three public log datasets from different systems. Experimental results show that FusionLog achieves over 90\% F1-score under a zero‑label setting, significantly outperforming state‑of‑the‑art cross‑system log‑based anomaly detection methods. 
\end{abstract}


\section{Introduction}
\IEEEPARstart{A}{s} the scale and complexity of software systems continue to grow, the frequency of failures has shown an upward trend. Ensuring the stability and reliability of systems has become one of the core challenges for their successful operation. System logs, which record key events and state changes, have become an essential source of information for anomaly detection~\cite{11229624, midlog, logcae, llmelog, zhao2025generality, 10.1145/3660800}. Log-based anomaly detection holds significant promise for enhancing system reliability and has emerged as a research hotspot in the current field.

Existing log-based anomaly detection models can mainly be divided into unsupervised and supervised models. Unsupervised models~\cite{10.1145/3377813.3381371, 9240683} use sequential neural networks to learn the occurrence probabilities of log events in normal event sequences, predicting subsequent log events and identifying events that deviate from the predictions as anomalies. However, due to the lack of explicit labeling of anomaly logs, the detection capability of these models is somewhat limited~\cite{9401970}. In contrast, supervised models~\cite{8854736, 10.1145/3338906.3338931} construct classification models to identify anomalous logs, typically demonstrating higher detection performance. However, their effectiveness largely depends on a large number of labeled logs. In real-world software systems, due to the fact that anomaly logs are often buried among a large amount of normal logs, obtaining accurate labels is a scarce and complex task~\cite{10.1145/3534678.3539106}. Therefore, applying supervised models to newly deployed software systems is highly challenging. 

To address the above issues, researchers have proposed cross‑system log‑based anomaly detection methods that borrow knowledge from mature systems via transfer learning ~\cite{9251092, 10.1145/3459637.3482209} or meta‑learning ~\cite{10.1145/3597503.3639205, FreeLog}. By transferring general knowledge from a mature system to a new system, these methods reduce reliance on large amounts of labeled logs. However, existing studies have shown that transfer learning methods guarantee performance only under specific assumptions and may face substantial difficulties when the distribution discrepancy is significant~\cite{ijcai2022p496}. As a result, the capability of these methods is heavily constrained. In contrast, meta-learning involves external optimization, enabling model to handle broader meta-representations beyond just model parameters~\cite{9428530}. Compared to transfer learning, meta-learning can achieve comparable generalization results with fewer logs ~\cite{gu-etal-2018-meta}. However, whether based on transfer learning or meta‑learning, existing studies concentrate on extracting general knowledge from a global perspective and overlook the significant discrepancies at the level of proprietary knowledge between the source and target systems.

\begin{figure*}[h!]
\centering
\includegraphics[width=2.0\columnwidth]{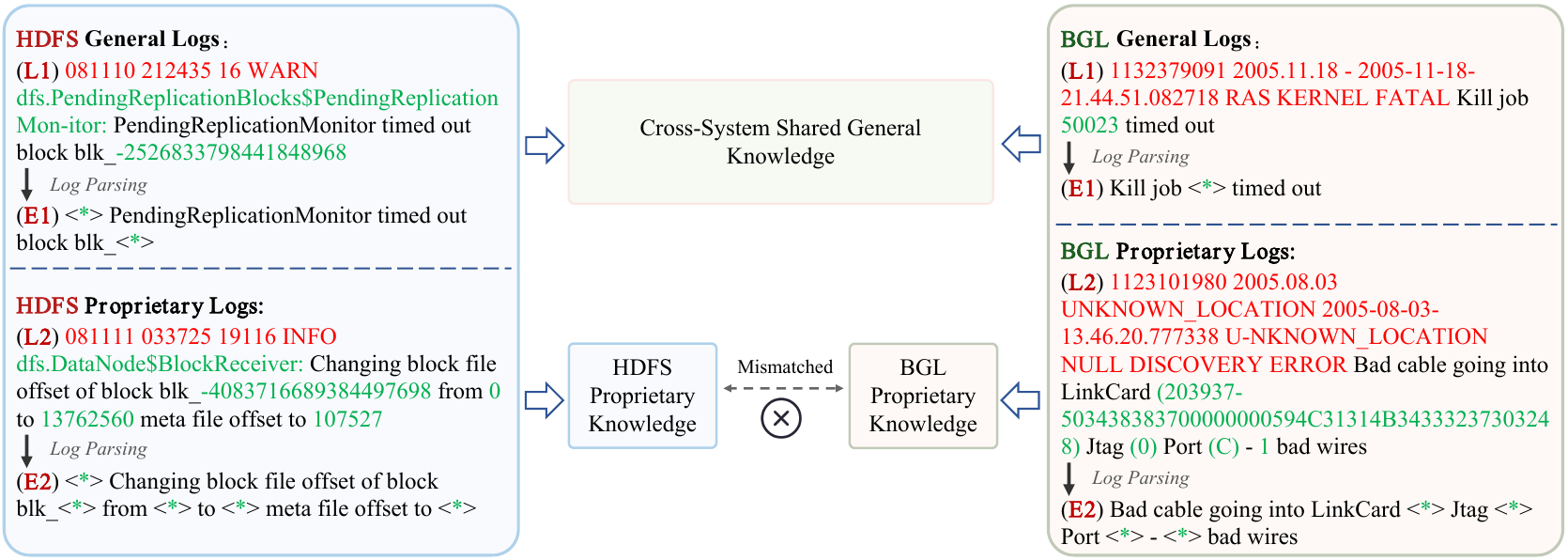}
\caption{Illustrative examples of general logs and proprietary logs from HDFS and BGL systems. General logs exhibit shared timeout-related patterns across systems, while proprietary logs capture system-specific events unique to each system.}
\label{fig_1}
\end{figure*}

Specifically, during system operation, logs are generated by the underlying code. Because different systems are maintained by different developers, logs often exhibit inconsistencies in naming, format, and terminology. Even when logs are semantically equivalent, their expressions may differ markedly across systems; moreover, because each system implements unique functionality, it produces system-specific logs that are absent in other systems. Through a preliminary study on the system logs of existing systems (detailed in Section~\ref{sec2}), we observe that cross-system logs can be classified into two categories: "general logs" that exhibit shared semantic patterns across systems, and "proprietary logs" that are unique to a specific system. Accordingly, general knowledge and proprietary knowledge are derived from these two categories, respectively.

Building on the above studies, although existing methods have achieved a certain success, their effectiveness is based on two mild assumptions: (1) that system events are shared across different systems. When the discrepancy in system events among systems becomes too great, their effectiveness cannot be guaranteed, resulting in unsatisfactory performance; (2) that the target system can supply a sufficient number of labeled logs, including enough proprietary logs. In the absence of labeled logs, the model cannot acquire the target system’s proprietary knowledge, and the extracted general knowledge is ineffective for proprietary logs, causing the model to fail to capture anomaly patterns unique to the target system. In summary, the pronounced disparity in proprietary knowledge across different systems, coupled with the mismatch between general and proprietary knowledge, constitutes a significant barrier to effective cross‑system log-based anomaly detection. This problem gives rise to two key technical challenges: (1) how to accurately distinguish and route different categories of logs; and (2) how to effectively fuse general and proprietary knowledge in an unlabeled setting.

To address these challenges, we propose FusionLog, a novel cross‑system log-based anomaly detection method based on the collaboration between LLM and small model. Specifically, to tackle the first challenge, we design a training-free router based on log semantic similarity that dynamically partitions the unlabeled target logs into “general logs” and “proprietary logs,” thereby providing fine‑grained inputs for subsequent processing. To address the second challenge, we design a dual-branch processing method. For general logs, we adopt a small model based on system‑agnostic representation meta‑learning for direct training and inference, inheriting general patterns between the source and target systems. For proprietary logs, we generate pseudo‑labels and fine‑tune the previously introduced small model via multi-round collaborative knowledge distillation and fusion based on LLM and small model to enhance its recognition of target-specific anomaly patterns. We evaluate the performance of FusionLog on three public log datasets from different systems (HDFS, BGL and OpenStack). Results show that under zero-label conditions, FusionLog achieves over 90\% F1-score, significantly outperforming state‑of‑the‑art cross‑system log-based anomaly detection methods. 

In summary, the contributions are as follows:
\begin{itemize}[leftmargin=*]
    \item{This work is the first to conceptualize cross‑system logs at the knowledge level by dividing them into “general logs” and “proprietary logs,” and to design specialized processing strategies based on their distinct characteristics. This breaks away from existing approaches that focus solely on general log patterns and ignore proprietary logs of the target system.}
    \item{We propose FusionLog, a novel cross‑system log‑based anomaly detection method, which effectively fuses general knowledge and proprietary knowledge via semantic routing and multi-round collaborative knowledge distillation and fusion based on LLM and small model.}
    \item{Evaluation results on three public log datasets demonstrate the significant effectiveness of our method.}
\end{itemize}

\begin{figure*}[h!]
    \centering
    \begin{minipage}{1.0\columnwidth}
        \centering
        \includegraphics[width=\linewidth]{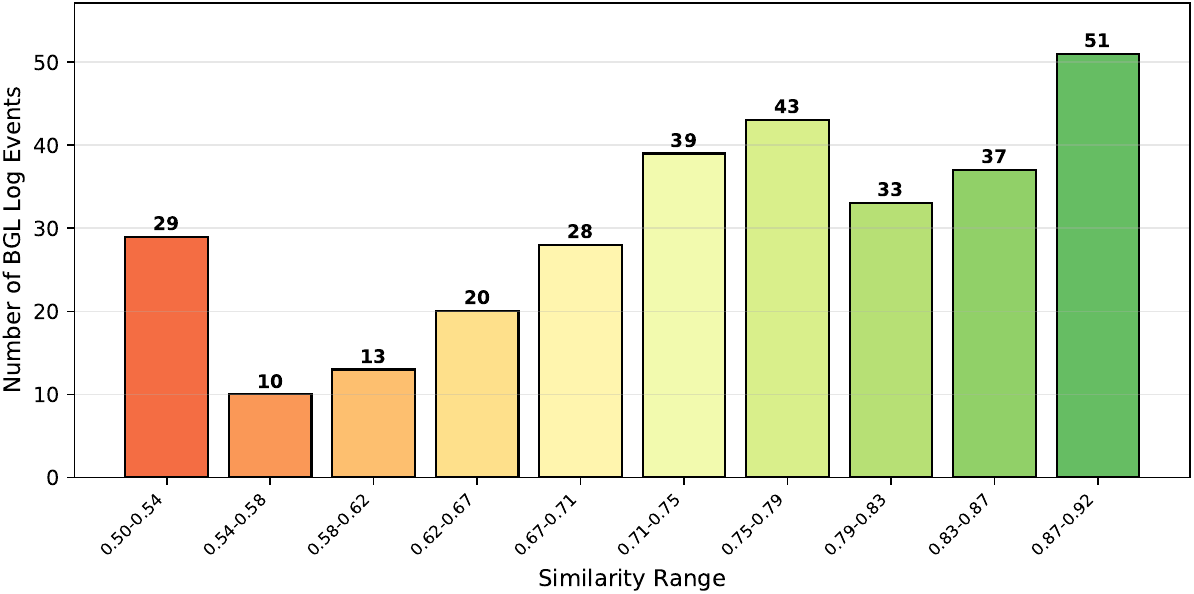}
        \caption{Max similarity distribution of BGL events to HDFS.}
        \label{fig_1.1}
    \end{minipage}
    \hfill
    \begin{minipage}{1.0\columnwidth}
        \centering
        \includegraphics[width=\linewidth]{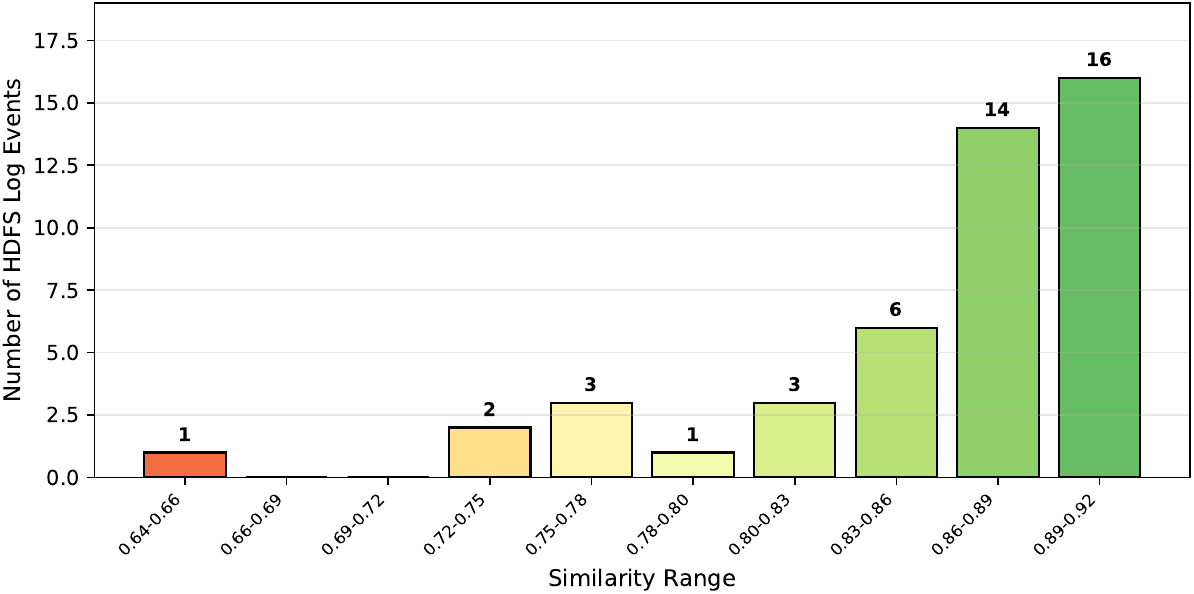}
        \caption{Max similarity distribution of HDFS events to BGL.}
        \label{fig_1.2}
    \end{minipage}
\end{figure*}

\section{Preliminary}
\label{sec2}
\subsection{System Logs and Log Events}
Software systems periodically record their operational status (e.g., parameters, execution state, events, etc.) in the form of text message sequences within logs. A log sequence consists of multiple log entries arranged chronologically. An event is an abstraction of a print statement in source code, which manifests itself in logs with different embedded parameter values in different executions—represented as a set of invariant keywords and parameters. An event can be used to summarize multiple log entries. An event sequence consists of a sequence of log events in one-to-one correspondence with log entries in a log sequence. Figure~\ref{fig_1} shows examples of logs and their corresponding log events from HDFS~\cite{10.1145/1629575.1629587} and BGL~\cite{4273008}. In the raw logs, the red part represents timestamps and log levels, and the green part represents variables, which are replaced with the placeholder "\textless*\textgreater".

\subsection{General Logs and Proprietary Logs}
Our observations of real-world logs reveal that there exist both "general logs" and "proprietary logs" between the source and target systems. For example, Figure \ref{fig_1} lists general logs and proprietary logs from two different systems. HDFS logs record file-system operations such as file access, data replication, and node status. These logs are primarily concerned with file system operations and are unrelated to hardware operations. BGL logs capture hardware status, task scheduling, and parallel job execution in a supercomputing environment. Specifically, the first two general logs are both associated with timeout events and describe similar failure conditions, illustrating shared operational patterns. The third entry records a DataNode block‑file offset change in the distributed file system, which is a proprietary event unique to HDFS with no corresponding record in BGL. Similarly, the fourth log reports a hardware‑level cable failure alert in the supercomputing system, a proprietary event that does not appear in HDFS logs. These examples demonstrate the disparity and mismatch between the cross‑system general knowledge and the target system’s proprietary knowledge. Such divergence hinders a model’s ability to learn the proprietary knowledge, thus constraining performance. Formally, general logs comprise entries that exhibit consistent semantic or structural characteristics across multiple systems, thereby reflecting shared operational patterns; in contrast, proprietary logs consist of entries unique to a particular system, characterized by distinct semantics or formats that capture its specific behaviors.

In addition, in Figures \ref{fig_1.1} and \ref{fig_1.2}, we statistically analyze the distribution of semantic similarities between log events in the HDFS and BGL datasets. HDFS contains 46 log event templates, while BGL contains 303 templates. Specifically, each log event in one system is matched to its most similar counterpart in the other system via maximum cosine similarity, with Figures \ref{fig_1.1} and \ref{fig_1.2} showing the BGL to HDFS and HDFS to BGL directions. The results show that both datasets contain log events with high similarity, corresponding to general logs, as well as events with low similarity, corresponding to proprietary logs, with each category occupying different proportions. This observation intuitively demonstrates the coexistence of general knowledge and proprietary knowledge, providing empirical support for the subsequent partitioning of logs into general and proprietary categories based on semantic similarity.

\begin{figure*}[h!]
\centering
\includegraphics[width=1.0\textwidth]{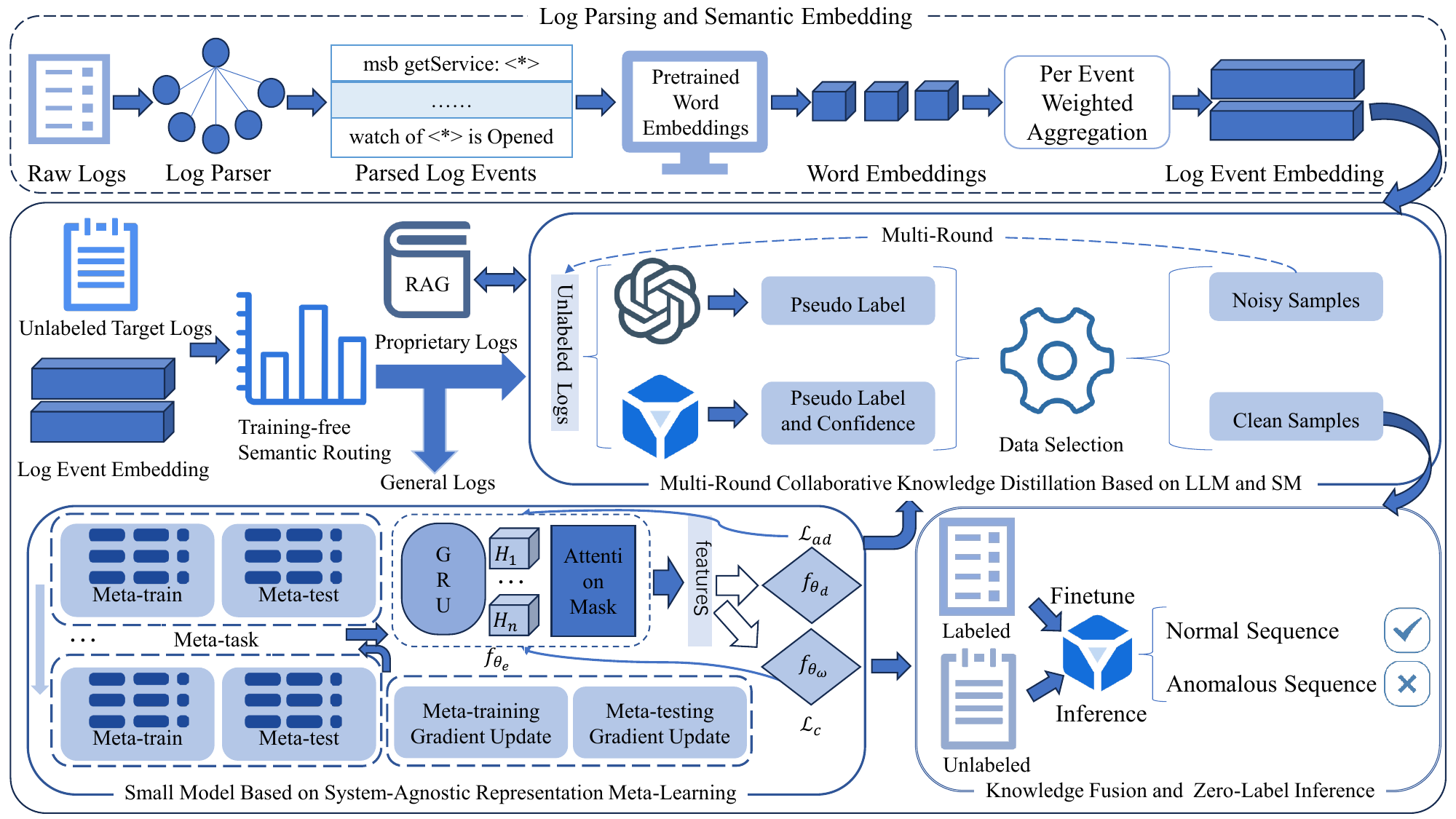}
\caption{Overview of FusionLog: a dual-phase pipeline integrating semantic routing, system-agnostic meta-learning, and LLM-SM collaborative knowledge distillation for cross-system log-based anomaly detection.}
\label{fig_2}
\end{figure*}

\section{Method}
\label{sec3}
\subsection{Overview}
To address the substantial mismatch between general knowledge and the proprietary knowledge of the target system, we propose FusionLog, a novel cross‑system log-based anomaly detection method. FusionLog comprises three core components: Training-free Semantic Routing, Small Model Based on System-Agnostic Representation Meta-Learning, and Multi-Round Collaborative Knowledge Distillation and Fusion Based on LLM and Small Model. Through their synergistic collaboration, FusionLog is able to concurrently capture both general knowledge and proprietary knowledge in a zero‑label scenario. Specifically, FusionLog operates in two sequential phases. In Phase I, unstructured logs are parsed into log events and embedded into semantic vectors, which the router uses to dynamically partition unlabeled target logs into "general logs" and "proprietary logs." In Phase II, general logs are processed by a small model based on system-agnostic representation meta-learning to inherit shared anomaly patterns, while proprietary logs are handled through multi-round collaborative knowledge distillation and fusion based on LLM and small model, which iteratively generates pseudo-labels and fine-tunes the small model to capture target-specific anomaly patterns. This dual‑branch architecture enables FusionLog to exploit general knowledge while accurately capturing proprietary knowledge. The complete workflow of FusionLog is illustrated in Figure \ref{fig_2}. The algorithm is provided in the Algorithm \hyperref[alg:fusionlog]{1}.

\subsection{Log Preprocessing and Semantic Routing}
\label{sec3.2}

FusionLog begins by employing the classic log parsing technique Drain~\cite{8029742} to process unstructured raw logs from various systems. Each raw log entry $l_i$ is parsed into a log event $e_i$, which replaces variable parameters with placeholders and retains only the invariant template keywords. Compared to traditional index-based methods, semantic embeddings have been shown to provide more informative representations. To account for the cross-system nature, we adopt the semantic embedding approach inspired by MetaLog~\cite{10.1145/3597503.3639205}, which constructs semantic embedding vectors for log events within a shared global space to ensure cross-system consistency. Specifically, the event text of $e_i$ is first tokenized, and each token $w_t$ is mapped into a $d$-dimensional vector using pre-trained GloVe word embeddings. Then, each token is weighted by its TF–IDF score, which combines the term frequency within the event and the inverse document frequency across both the source and target log corpora. The weighted word vectors are summed to produce the semantic embedding vector ${v}_i \in \mathbb{R}^d$ for event $e_i$. Through this process, each log entry $l_i$ is associated with its event embedding ${v}_i$ for subsequent routing.

After obtaining semantic embeddings for each log event, FusionLog then performs semantic routing to partition the unlabeled target log sequences into "general logs" and "proprietary logs." The procedure is as follows: Let a target log sequence be $x_k = \{l_1, l_2, \dots, l_n\}$, where each log entry $l_i$ has an event embedding vector ${v}_i \in \mathbb{R}^d$. Denote the set of all event embeddings of the source system by ${V}^{source} = \{u_1, u_2, \dots, u_m\}, u_j \in \mathbb{R}^d$. Compute the cosine similarity between ${v}_i$ and each ${u}_j$, and take the maximum:
\[
sim_i = \max_{1 \le j \le m} \mathrm{cosine}({v}_i, {u}_j) = \max_{j} \frac{{v}_i \cdot {u}_j}{\|{v}_i\|\;\|{u}_j\|}
\]
To assess the overall similarity of $x_k$ to the source domain, we aggregate by taking the minimum event‑level score:
\[
{x_k}_{\mathrm{sim}} = \min_{1 \le i \le n} sim_i
\]
This highlights the event that is least similar to the source domain (i.e., the bottleneck event), and since $|{V}^{source}|$ is only on the order of tens or hundreds, it keeps the computation efficient. Given a threshold $\tau \in [0,1]$, the sample $x_k$ is routed to the general logs when its similarity score ${x_k}_{\mathrm{sim}}$ exceeds $\tau$; otherwise, it is routed to the proprietary logs. In practice, we set $\tau$ to the mean of all sequence-level similarity scores, as this provides a natural partition point that adapts to the similarity distribution of each system pair without requiring manual tuning. A detailed sensitivity analysis of $\tau$ is provided in Section~\ref{sec4.4}.

The rationale for semantic routing is as follows. We adopt minimum rather than mean or median aggregation because a single proprietary event within an otherwise general sequence is sufficient to indicate that the sequence requires specialized processing. Moreover, the low similarity of proprietary events reflects genuine semantic divergence rather than mere rarity: as illustrated in Figure \ref{fig_1}, proprietary events such as HDFS's DataNode operations and BGL's hardware cable alerts represent fundamentally different system functionalities with no semantic counterpart in the other system. It is also important to note that the partition is an operational distinction rather than a rigid taxonomy. FusionLog's dual-branch architecture is inherently tolerant of routing imprecision: misrouted samples are still processed by the alternative branch without catastrophic failure, resulting only in suboptimal processing for a small fraction of data. This robustness is empirically validated by the threshold sensitivity analysis in Section~\ref{sec4.4}.

\subsection{Small Model Based on System-Agnostic Representation Meta-Learning}
\label{sec3.3}

For general logs, the small model is able to effectively extract and generalize the general knowledge. FusionLog adopts a small model based on system‑agnostic representation meta‑learning for direct training and inference ~\cite{FreeLog}. The small model of FusionLog consists of two key stages: adversarial unsupervised domain adaptation and meta-learning. Specifically, we use \( X_S \) and \( X_T \) to represent the logs sampled from the source domain \( D_S \) and the target domain \( D_T \), while \( Y_S \) denotes the label matrix for \( X_S \). We first construct a cross-system meta-task \( MT_i = \{ M_i^{sup}, M_i^{que} \}\), where \( M_i^{sup} = \{ X_{S_i}^{sup}, X_{T_i}^{sup}, Y_{S_i}^{sup} \} \) is used for meta-training, \( M_i^{que} = \{ X_{S_i}^{que}, X_{T_i}^{que}, Y_{S_i}^{que} \} \) is used for meta-testing, and \( X^{sup} \) and \( X^{que} \) are referred to as the support and query set. 

The small model of FusionLog consists of three main modules: the feature extractor \( f_{\theta_e} \), the anomaly classifier \( f_{\theta_\omega} \) and the domain classifier \( f_{\theta_d} \). Like MetaLog~\cite{10.1145/3597503.3639205} and FreeLog~\cite{FreeLog}, \( f_{\theta_e} \) consists of two modules: the Gated Recurrent Unit (GRU) and the attention mask layer. Given a chronologically ordered sequence of log event embeddings, the GRU module maintains a hidden state at each time step, modeling temporal dependencies and sequential ordering among log events. The attention module then takes the hidden states as input and utilizes adaptive self-attention to fuse the information into a final sequence representation. The output of \( f_{\theta_e} \) is fed into both the anomaly classifier \( f_{\theta_\omega} \), which generates an anomaly probability, and the domain classifier \( f_{\theta_d} \), which outputs a domain classification result.

Specifically, in each meta-task, given the feature extractor \( f_{\theta_e} \), we train the domain classifier \( f_{\theta_d} \) to maximize the distinction between features from the source and target domains. The optimization problem is as follows:
\[
\max_{\theta_d} \sum_{MT_i} {L}_{ad}^{MT_i}(M_i^{sup}; f_{\theta_d})
\]
where the adversarial loss function is the binary cross-entropy with logits loss. The update of \( f_{\theta_d} \) can be written as:
\[
\theta_d \gets \theta_d + \lambda \nabla_{\theta_d} \sum_{MT_i} {L}_{ad}^{MT_i}(M_i^{sup}; f_{\theta_d})
\]
Then, we train the anomaly classifier \( f_{\theta_\omega} \) to learn discriminative features for classifying normal and anomalous logs. The optimization problem is as follows:
\[
\min_{\theta_\omega} \sum_{MT_i} {L}_c^{MT_i}(M_i^{sup}; f_{\theta_\omega})
\]
where the classification loss function is the binary cross-entropy loss. The update of \( f_{\theta_\omega} \) can be written as:
\[
\theta_\omega \gets \theta_\omega - \kappa \nabla_{\theta_\omega} \sum_{MT_i} {L}_c^{MT_i}(M_i^{sup}; f_{\theta_\omega})
\]
where $\kappa$ and $\lambda$ denote the learning rates.

During the meta-training phase, the learner's parameters can be updated through one or more gradient descent steps:
\[
\theta_e^i = \theta_e - \delta \nabla_{\theta_e} {L}_{MT_i}(M_i^{sup}; f_{\theta_e})
\]
where \( \delta \) is the learning rate and the objective function is:
\[
\resizebox{0.98\columnwidth}{!}{$
\mathcal{L}_{MT_i}(f_{\theta_e}) = \gamma \mathcal{L}_c^{MT_i}(X_{S_i}^{sup}, Y_{S_i}^{sup}; f_{\theta_\omega}) - \beta \mathcal{L}_{ad}^{MT_i}(X_{S_i}^{sup}, X_{T_i}^{sup}; f_{\theta_d})
$}
\]
The first term represents the classification loss in the source domain. The second term is the domain adversarial loss, which encourages \( f_{\theta_e} \) to produce domain-invariant features by aligning the domains through \( f_{\theta_d} \). The hyperparameters \( \beta \) and \( \gamma \) control the trade-off between adaptation and classification performance. This method integrates classification loss and adversarial loss, enabling the model to effectively generalize from the source system to the target system. 

After learning the adaptation parameters \( \theta_e^i \) for each task, we proceed to meta-optimize the feature extractor \( f_{\theta_e} \) to improve the performance of \( \theta_e^i \) on the query set. The meta-objective function can be expressed as:
\[
\min_{\theta_e} \sum_{MT_i} {L}_{MT_i}(M_i^{que}; f_{\theta_e^i})
\]
We perform meta-optimization via gradient descent as follows:
\[
\theta_e \gets \theta_e - \alpha \nabla_{\theta_e} \sum_{MT_i} {L}_{MT_i}(M_i^{que}; f_{\theta_e^i})
\]
where \( \alpha \) is the meta-step size.

Overall, in each meta-task, the model performs adversarial training for unsupervised domain adaptation to extract general knowledge between labeled source logs and unlabeled target logs. At the same time, the model computes classification loss using the labeled source logs to extract discriminative features related to anomaly detection. Through the dual optimization of adversarial domain alignment and classification discrimination, the feature extractor is updated in the support set, while the initialization parameters are reverse optimized in the query set, enabling the model to maintain both classification power and domain alignment ability without labeled training.

\subsection{Multi-Round Collaborative Knowledge Distillation and Fusion Based on Large Language Model and Small Model}
\label{sec3.4}

For proprietary logs, FusionLog employs a multi‑round collaborative knowledge distillation and fusion approach that combines LLM with small model. By leveraging the LLM’s powerful comprehension capabilities to produce high‑quality pseudo‑labels for unlabeled logs, together with the small model’s lightweight inference and rapid adaptation, we iteratively filter “clean” samples and use them both to fine‑tune the previously introduced small model and to enrich the LLM’s RAG knowledge base, thereby progressively enhancing the model’s ability to detect proprietary anomalies ~\cite{Zhou_Zhang_Tan_Zhang_Li_2025}.

In the first iteration, for each proprietary log sequence $x_i$, we retrieve a set of highly relevant in‑context examples ${D}'^{(r)}$ from the RAG knowledge base ${K}^{(r)}$, which is initially constructed by applying the trained small model to the general log subset, rather than the proprietary subset, because at this stage the small model produces reliable predictions only on general logs (F1-score above 93\% as shown in Figure \ref{fig_3}) while its accuracy on proprietary logs remains low (F1-score of 54\%). This ensures that the initial knowledge base is predominantly composed of correctly labeled examples, providing the LLM with trustworthy in-context references. These examples are inserted into the LLM's prompt to generate a pseudo‑label:
\[
\hat y_i^{\mathrm{LLM}} = \mathrm{LLM}(x_i;\,{K}^{(r)},\,{D}'^{(r)})
\]
Concurrently, the small model ${SM}$ performs a forward pass on the same sample to produce its own pseudo‑label $\hat y_i^{\mathrm{SM}}$ and an associated confidence score $p(\hat y_i^{\mathrm{SM}})$. The data selection module then partitions the logs into "clean" and "noisy" subsets based on label agreement and a confidence threshold $\epsilon$: samples enter the clean pool $D_{\mathrm{clean}}^{(r)}$ if $\hat y_i^{\mathrm{LLM}} = \hat y_i^{\mathrm{SM}}$ and $p(\hat y_i^{\mathrm{SM}}) \ge \epsilon$, otherwise they enter the noisy pool $D_{\mathrm{noisy}}^{(r)}$. By comparing the pseudo-labels generated by the LLM with those produced by the small model, and applying confidence-based filtering, potential erroneous pseudo-labels from the LLM can be effectively filtered out, thereby mitigating error propagation and enhancing the stability and reliability of the knowledge distillation process. It is noteworthy that as the number of distillation rounds increases, the small model's performance continuously improves, and the corresponding threshold $\epsilon$ dynamically decreases to align with its ongoing optimization. Compared to the static threshold scheme, this dynamic threshold strategy significantly enhances the efficiency of knowledge distillation by preventing training stagnation that can occur when threshold adjustments lag behind model performance gains. Specifically, we set the initial threshold $\epsilon_0 = 0.9$ to ensure that only high-confidence, mutually agreed-upon samples enter the clean pool in the first round, when the small model is least reliable. The threshold then decreases linearly by a fixed step of 0.05 per round, gradually relaxing the filtering criterion as the small model is progressively fine-tuned and becomes more capable of producing accurate predictions. This linear schedule is chosen for its simplicity and predictability; we empirically verify its effectiveness against alternative configurations in the parameter sensitivity analysis (Section~\ref{sec4.4}). For a comparison between dynamic and static thresholds, see Figure \ref{fig_4} and the ablation study section.

Each round's clean dataset $D_{\mathrm{clean}}^{(r)}$ is then used to fine‑tune the small model, strengthening its discrimination of proprietary anomaly patterns:
\[
\theta_{{SM}}^{(r+1)} \leftarrow \mathrm{FineTune}(\theta_{{SM}}^{(r)},\,D_{\mathrm{clean}}^{(r)})
\]
Simultaneously, these high‑quality samples are merged into the RAG knowledge base to augment the LLM's next iteration of retrieval:
\[
{K}^{(r+1)} = {K}^{(r)} \cup D_{\mathrm{clean}}^{(r)}
\]
After updating, the process repeats on the remaining noisy pool: pseudo‑label generation, data selection, and small model fine‑tuning are applied in successive iterations, continually expanding the clean dataset and improving the joint labeling accuracy. This optimization loop continues until the preset maximum number of iterations ${N}$ is reached. In the final iteration, samples that remain in the noisy pool $D_{\mathrm{noisy}}^{({N})}$ typically correspond to low-frequency events in the target logs. Since these samples cannot be reliably filtered by the small model, their labels are directly taken from the predictions of the LLM, ensuring that the proprietary knowledge distillation process covers all event types in the target system.

Through this collaborative distillation and iterative refinement, FusionLog progressively filters and integrates proprietary knowledge into the small model in a fully unlabeled environment. Regarding the choice of the maximum iteration count $N$, we observe that the noisy pool naturally shrinks as distillation progresses, since each round converts a portion of noisy samples into clean samples. As shown in Figure \ref{fig_4}, the noisy pool ratio declines from 55\% in Round 1 to 0\% by Round 5, at which point all proprietary logs have been processed, indicating that $N=5$ is the empirically observed number of rounds required to exhaust the noisy pool for our datasets. Since each subsequent round only processes the diminishing noisy pool, the computational cost of later rounds is negligible. In practice, an early stopping criterion based on noisy pool size could be adopted to reduce unnecessary iterations when convergence occurs sooner.

After the training phase, the fused small model $\theta_{SM}^{(N)}$ has absorbed both general knowledge (via meta-learning on the source and target general logs) and proprietary knowledge (via collaborative distillation on the target proprietary logs). At inference time, all incoming target log sequences are fed directly into this single fused small model for end-to-end anomaly prediction, without requiring semantic routing, LLM invocation, or RAG retrieval.

\begin{algorithm}[h!]
\caption{FusionLog}
\label{alg:fusionlog}
\renewcommand{\algorithmicrequire}{\textbf{Input:}}
\renewcommand{\algorithmicensure}{\textbf{Output:}}
\begin{algorithmic}[1]
\REQUIRE Source logs $D_S$, target logs $D_T$; task distribution $p(MT)$; max rounds $N$; learning rates $\delta,\lambda,\kappa,\alpha$; routing threshold $\tau$; initial confidence threshold $\epsilon_0$; hyperparameters $\beta,\gamma$
\ENSURE Trained small model $\theta_{SM}^{(N)}$

\STATE \textbf{--- Phase I: Log Preprocessing \& Semantic Routing ---}
\STATE Parse $D_S, D_T$ via Drain; embed each event $l_i \!\to\! v_i$
\FORALL{target sequence $x_k = \{l_1, \dots, l_n\} \in D_T$}
    \STATE $s_k \gets \min_{1 \le i \le n}\, \max_{u_j \in V^{source}} \operatorname{cosine}(v_i, u_j)$
    \STATE Route $x_k$ to $\mathcal{G}$ (general) if $s_k \ge \tau$, else to $\mathcal{P}$ (proprietary)
\ENDFOR

\STATE \textbf{--- Phase II-A: Meta-Learning on General Logs ---}
\STATE Randomly initialize $\theta_e, \theta_\omega, \theta_d$
\WHILE{not converged}
    \STATE Sample meta-task batch $\{MT_i\} \!\sim\! p(MT)$, $MT_i \!=\! (M_i^{sup}, M_i^{que})$
    \STATE $\theta_d \!\gets\! \theta_d + \lambda \nabla_{\theta_d} \sum_i L_{ad}^{MT_i}(M_i^{sup})$
    \STATE $\theta_\omega \!\gets\! \theta_\omega - \kappa \nabla_{\theta_\omega} \sum_i L_c^{MT_i}(M_i^{sup})$
    \FORALL{$MT_i$}
        \STATE $\theta_e^i \!\gets\! \theta_e - \delta \nabla_{\theta_e} L_{MT_i}(M_i^{sup})$
    \ENDFOR
    \STATE $\theta_e \!\gets\! \theta_e - \alpha \nabla_{\theta_e} \sum_i L_{MT_i}(M_i^{que};\, f_{\theta_e^i})$
\ENDWHILE

\STATE \textbf{--- Phase II-B: Collaborative Distillation on Proprietary Logs ---}
\STATE $\theta_{SM}^{(0)} \!\leftarrow\! (\theta_e, \theta_\omega, \theta_d)$
\STATE Build $K^{(0)}$ from $\mathcal{G}$ using SM predictions and confidence
\FOR{$r = 0$ \TO $N\!-\!1$}
    \STATE $\epsilon^{(r)} \gets$ dynamically decreased from $\epsilon_0$
    \STATE $D_{\mathrm{clean}}^{(r)},\, D_{\mathrm{noisy}}^{(r)} \leftarrow \emptyset$
    \FORALL{$x_i \in \mathcal{P} \setminus \bigcup_{t<r} D_{\mathrm{clean}}^{(t)}$}
        \STATE Retrieve $D'^{(r)}$ from $K^{(r)}$
        \STATE $\hat{y}_i^{\mathrm{LLM}} \!\gets\! \mathrm{LLM}(x_i;\, K^{(r)}, D'^{(r)})$
        \STATE $(\hat{y}_i^{\mathrm{SM}},\, p_i) \!\gets\! SM_{\theta_{SM}^{(r)}}(x_i)$
        \IF{$\hat{y}_i^{\mathrm{LLM}} \!=\! \hat{y}_i^{\mathrm{SM}}$ \textbf{and} $p_i \!\ge\! \epsilon^{(r)}$}
            \STATE $D_{\mathrm{clean}}^{(r)} \!\leftarrow\! D_{\mathrm{clean}}^{(r)} \cup \{(x_i,\, \hat{y}_i^{\mathrm{LLM}})\}$
        \ELSE
            \STATE $D_{\mathrm{noisy}}^{(r)} \!\leftarrow\! D_{\mathrm{noisy}}^{(r)} \cup \{x_i\}$
        \ENDIF
    \ENDFOR
    \STATE $\theta_{SM}^{(r+1)} \!\gets\! \mathrm{FineTune}(\theta_{SM}^{(r)},\, D_{\mathrm{clean}}^{(r)})$
    \STATE $K^{(r+1)} \!\gets\! K^{(r)} \cup D_{\mathrm{clean}}^{(r)}$
\ENDFOR
\STATE Label remaining $D_{\mathrm{noisy}}^{(N-1)}$ directly with LLM predictions
\STATE Fine-tune $\theta_{SM}^{(N)}$ on $D_{\mathrm{noisy}}^{(N-1)}$ with LLM labels

\RETURN $\theta_{SM}^{(N)}$
\end{algorithmic}
\end{algorithm}

\section{Experiments}
\label{sec4}
\subsection{Experimental Setup}
\label{sec4.1}

\textbf{\textit{Datasets.}} We conducted experiments on three publicly available log datasets: HDFS~\cite{10.1145/1629575.1629587}, BGL~\cite{4273008}, and OpenStack~\cite{10.1145/3133956.3134015}.
\begin{itemize}[leftmargin=*]
    \item \textbf{HDFS} logs record file-system operations in the Hadoop Distributed File System. Anomalies are labeled via handcrafted rules, and logs are grouped into traces by block IDs.
    \item \textbf{BGL} logs are collected from the BlueGene/L supercomputer at Lawrence Livermore National Laboratory, capturing hardware status, task scheduling, and parallel execution, with messages tagged by alert category.
    \item \textbf{OpenStack} logs are generated on CloudLab with injected failures, recording operations and errors across cloud infrastructure components.
\end{itemize}

These three systems span distinctly different operational domains, including distributed file storage (HDFS), high-performance computing (BGL), and cloud infrastructure management (OpenStack), representing diverse real-world cross-system scenarios. The dataset statistics are summarized in Table \hyperref[tab5]{1}.

\begin{table}[h!]
\caption{Statistics of the Datasets}
\centering
\resizebox{1.0\columnwidth}{!}{
\begin{tabular}{lcccc}
\toprule
\textbf{Dataset} & \textbf{Lines} & \textbf{Sequences} & \textbf{Normal} & \textbf{Anomalous} \\
\midrule
HDFS~\cite{10.1145/1629575.1629587} & 11,175,629 & 575,061 & 558,223 & 16,838 \\
BGL~\cite{4273008} & 4,747,963 & 85,576 & 49,273 & 36,303 \\
OpenStack~\cite{10.1145/3133956.3134015} & 207,820 & 9,897 & 9,020 & 877 \\
\bottomrule
\end{tabular}}
\label{tab5}
\end{table}

We selected four cross-system combinations (HDFS to BGL, BGL to HDFS, OpenStack to HDFS and OpenStack to BGL) for evaluation. Following~\cite{9401970}, we used Drain~\cite{8029742} to parse log events and organize log sequences, ensuring consistency with prior methods. For HDFS, log sequences are grouped by block IDs; for BGL and OpenStack, a fixed-window strategy is used to group consecutive log entries within a predefined time window into sequences.

\textbf{\textit{Baselines.}} To ensure a fair comparison with FusionLog, we adopted various baseline methods and experimental setups. Due to the significant differences in the sizes of the three datasets, we used different proportions of data for the experiments, as shown in Table \hyperref[tab_1]{2}. All data proportions and configurations for the baselines follow the original experimental settings reported in their respective papers~\cite{9401970, 10.1145/3597503.3639205, FreeLog}, ensuring reproducibility and consistency with prior work. The baselines are organized into the following groups:

\begin{itemize}[leftmargin=*]

\item \textbf{Blocks (a) and (b): Target-system supervised baselines (PLELog~\cite{9401970}, LogRobust~\cite{10.1145/3338906.3338931}).}
Both blocks evaluate PLELog and LogRobust trained exclusively on the target system under different label availability:
\begin{itemize}
    \item \textit{Block (a) — Scarce anomaly labels:} When BGL is the target system, both methods (a1, a2) are trained on 30\% of normal log sequences and only 1\% of anomalous log sequences from BGL. When HDFS is the target system, they are trained on 10\% of normal log sequences and 1\% of anomalous log sequences from HDFS. This block evaluates the performance under a scenario where anomaly labels are scarce.
    \item \textit{Block (b) — Full labels:} When BGL is the target system, both methods (b1, b2) are trained on 30\% of total log sequences from BGL. When HDFS is the target system, they are trained on 10\% of total log sequences from HDFS. This block examines the capability with fully annotated (100\%) anomaly labels.
\end{itemize}

\item \textbf{Block (c): Transfer learning baselines (LogTAD~\cite{10.1145/3459637.3482209}, LogTransfer~\cite{9251092}, LogDLR~\cite{10910212}).}
These methods leverage partially labeled target logs and source logs:
\begin{itemize}
    \item \textit{(c1) LogTAD and (c2) LogTransfer:} In HDFS$\to$BGL, trained on 30\% of normal and 1\% of anomalous log sequences from BGL, together with 30\% of log sequences from HDFS. In BGL$\to$HDFS, trained on 10\% of normal and 1\% of anomalous log sequences from HDFS, together with all log sequences from BGL. OpenStack$\to$HDFS and OpenStack$\to$BGL follow the same target-system configurations with all log sequences from OpenStack.
    \item \textit{(c3) LogDLR:} Randomly selects 100,000 and 10,000 normal log sequences from the source and target system respectively for training.
\end{itemize}

\item \textbf{Block (d): Meta-learning baseline (MetaLog~\cite{10.1145/3597503.3639205}).}
MetaLog is evaluated under three settings:
\begin{itemize}
    \item \textit{(d1):} Shares the same source and target data configurations as LogTAD and LogTransfer.
    \item \textit{(d2):} Built on the (d1) configuration but with all anomaly labels removed, performing anomaly detection using only normal labels under the same settings.
    \item \textit{(d3):} Based on the FusionLog experimental configuration, with all anomaly labels removed, performing anomaly detection using only normal labels under the same settings.
\end{itemize}
Blocks (c) and (d) evaluate cross-system methods, based on prior transfer-learning and meta-learning approaches, that leverage partially labeled target logs.

\item \textbf{Block (e): Unsupervised baseline (DeepLog~\cite{10.1145/3133956.3134015}).}
DeepLog (e1) is trained exclusively on normal labels from the target dataset, maintaining consistency with other baseline methods.

\begin{table*}[h!]
\caption{Performance comparison of FusionLog and baselines on cross-system log-based anomaly detection.}
\centering
\label{tab_1}

\resizebox{2.0\columnwidth}{!}{
\begin{tabular}{lcccccccccccc}
\toprule
\multicolumn{1}{c}{\multirow{2}{*}{\textbf{Method}}} & \multicolumn{3}{c}{\textbf{HDFS to BGL}} & \multicolumn{3}{c}{\textbf{BGL to HDFS}} & \multicolumn{3}{c}{\textbf{OpenStack to HDFS}} & \multicolumn{3}{c}{\textbf{OpenStack to BGL}} \\
\cmidrule(lr){2-4} \cmidrule(lr){5-7} \cmidrule(lr){8-10} \cmidrule(lr){11-13}
 & \textbf{Precision} & \textbf{Recall} & \textbf{F1-score} & \textbf{Precision} & \textbf{Recall} & \textbf{F1-score} & \textbf{Precision} & \textbf{Recall} & \textbf{F1-score} & \textbf{Precision} & \textbf{Recall} & \textbf{F1-score} \\
\midrule
PLELog (a1)  & 82.10 & 67.42 & 74.04 & 65.86 & 71.11 & 68.38 & 65.86 & 71.11 & 68.38 & 82.10 & 67.42 & 74.04 \\
LogRobust (a2)  & 94.60 & 72.95 & 82.38 & 100.00 & 62.30 & 76.77 & 100.00 & 62.30 & 76.77 & 94.60 & 72.95 & 82.38 \\ 
\midrule
PLELog (b1)  & 94.88 & 89.62 & 92.18 & 96.30& 83.81& 89.62 & 96.30& 83.81& 89.62 & 94.88 & 89.62 & 92.18 \\
LogRobust (b2)  & 97.52 & 91.27 & 94.29 & 82.54& 99.20& 90.11 & 82.54& 99.20& 90.11 & 97.52 & 91.27 & 94.29 \\ 
\midrule
LogTAD (c1)  & 78.01 & 68.51 & 72.95 & 78.80 & 71.22 & 74.82 & 71.89 & 65.51 & 68.55 & 70.72 & 65.51 & 68.02 \\
LogTransfer (c2)  & 74.42 & 76.73 & 75.56 & 100.00 & 43.30 & 60.43 & 73.87 & 62.30 & 67.59 & 68.43 & 71.32 & 69.85 \\ 
LogDLR (c3)  & 79.25 & 72.56 & 75.76 & 77.78 & 70.05 & 73.71 & 73.65 & 69.80 & 71.67 & 69.58 & 64.33 & 66.85 \\ 
\midrule
MetaLog (d1)  & 96.89 & 89.28 & 92.93 & 89.29 & 74.98 & 81.51 & 96.67 & 62.42 & 75.86 & 99.83 & 70.09 & 82.36 \\
MetaLog (d2)  & 64.80 & 3.62 & 6.86 & 99.93 & 28.09 & 43.85 & 97.46 & 19.21 & 32.09 & 100.00 & 1.70 & 3.34 \\
MetaLog (d3)  & 98.75 & 20.05 & 33.33 & 72.29 & 12.10 & 20.73 & 100.00 & 18.35 & 31.01 & 100.00 & 1.20 & 2.37 \\  
\midrule
DeepLog (e1)  & 66.13 & 48.79 & 56.16 & 53.96 & 34.07 & 41.77 & 53.96 & 34.07 & 41.77 & 66.13 & 48.79 & 56.16 \\ 
\midrule
PLELog (f1)  & 38.80 & 99.87 & 55.89 & 1.69 & 92.85 & 3.32 & 4.33 & 53.47 & 8.01 & 54.65 & 43.04 & 48.16 \\
LogRobust (f2)  & 39.08 & 93.67 & 55.15 & 2.25 & 62.12 & 4.35 & 0.63 & 60.81 & 1.25 & 34.34 & 57.39 & 42.97 \\
NeuralLog (f3)  & 57.23 & 52.79 & 54.38 & 33.13 & 58.04 & 42.06 & 3.33 & 42.99 & 6.17 & 14.76 & 81.45 & 24.99 \\
MetaLog (f4)  & 29.43 & 0.45 & 0.89 & 2.90 & 80.92 & 5.61 & 2.90 & 80.92 & 5.61 & 29.43 & 0.45 & 0.89 \\
FreeLog (f5)  & 81.12 & 77.10 & 79.06 & 79.34 & 76.11 & 77.69 & 71.34 & 79.09 & 75.02 & 73.59 & 80.33 & 76.81 \\ 
\midrule
RAGLog (g1)  & 81.25 & 98.50 & 89.05 & 85.33 & 97.65 & 91.08 & 85.33 & 97.65 & 91.08 & 81.25 & 98.50 & 89.05 \\
RAGLog (g2)  & 45.76 & 100.00 & 62.79 & 49.50 & 100.00 & 66.22 & 49.50 & 100.00 & 66.22 & 45.76 & 100.00 & 62.79 \\
\midrule
\rowcolor{gray!15} Ours FusionLog & \textbf{95.21} & \textbf{94.10} & \textbf{94.65} & \textbf{95.58} & \textbf{90.73} & \textbf{93.09} & \textbf{91.78} & \textbf{93.75} & \textbf{92.76} & \textbf{93.94} & \textbf{95.45} & \textbf{94.69} \\
\bottomrule
\end{tabular}
}
\end{table*}

\item \textbf{Block (f): Zero-label baselines (PLELog, LogRobust, NeuralLog~\cite{9678773}, MetaLog, FreeLog~\cite{FreeLog}).}
These methods are evaluated under the zero-label setting where no target-system labels are available:
\begin{itemize}
    \item \textit{(f1) PLELog, (f2) LogRobust, (f3) NeuralLog:} Trained solely on subsets of the source datasets (all log sequences of BGL and OpenStack, 30\% of log sequences of HDFS) and directly tested on the target datasets.
    \item \textit{(f4) MetaLog:} Trained on the fully labeled logs of the two non-target systems and evaluated directly on the target system. When HDFS is the target, trained on 30\% of BGL log sequences and all OpenStack log sequences; when BGL is the target, trained on 10\% of HDFS log sequences and all OpenStack log sequences.
    \item \textit{(f5) FreeLog:} Labeled logs from the source system are used for training, while target-system logs remain unlabeled. In HDFS$\to$BGL, all anomalous log sequences from HDFS and an equal number of normal log sequences, along with half unlabeled logs from BGL, are used for training. BGL$\to$HDFS follows the same settings. For OpenStack$\to$HDFS and OpenStack$\to$BGL, since OpenStack is much smaller, all log sequences from OpenStack are used, along with an equal number of normal and anomalous half unlabeled logs from the target datasets.
\end{itemize}
Blocks (e) and (f) assess the performance of existing methods in the zero-label setting.

\item \textbf{Block (g): LLM-based baseline (RAGLog~\cite{10607047}).}
\begin{itemize}
    \item \textit{(g1):} Uses 10\% of the labeled logs from the target system as the RAG knowledge base.
    \item \textit{(g2):} Tests the LLM performance without using RAG that incorporates labeled logs from the target system.
\end{itemize}

\item \textbf{FusionLog (Ours).} We adopt the same experimental settings as FreeLog.

\end{itemize}

\textbf{\textit{Definition of Evaluation Metrics.}}
We selected precision, recall and F1-score as evaluation metrics, defined as follows:  
\(
\text{Precision} = \frac{TP}{TP + FP}, \quad \text{Recall} = \frac{TP}{TP + FN}, \quad F_1 = \frac{2 \cdot \text{Precision} \cdot \text{Recall}}{\text{Precision} + \text{Recall}},
\)
where \( TP \), \( FP \), and \( FN \) represent true positives, false positives, and false negatives, respectively.

\textbf{\textit{Implementation Details.}} The small model was trained on a single NVIDIA 3090 GPU using Adam optimizer with a batch size of 256 and a learning rate of 1e-3. Log event embeddings are 300-dimensional vectors following~\cite{9401970}. For the LLM, we used Qwen3-Plus~\cite{yang2025qwen3technicalreport} with temperature 1.0 and top\_p 0.8. The routing threshold $\tau$ is set to the mean of sequence-level similarity scores. The data selection threshold is initialized at $\epsilon_0=0.9$ and decreases by 0.05 per round over $N=5$ distillation rounds. During RAG retrieval, the top three most similar log sequences are selected as in-context examples, each containing the log content, the small model's predicted label, and confidence score, serving as reference for the LLM rather than merely providing labels. All datasets are accessible via Loghub~\cite{loghub1, loghub}, and the code is available at \href{https://github.com/XinlongZhao/FusionLog}{https://github.com/XinlongZhao/FusionLog}.

\subsection{Main Results}
Cross‑system generalization is an extremely challenging scenario in log-based anomaly detection. Table \hyperref[tab_1]{2} shows the comparative results of FusionLog and various baseline methods under four experimental settings. As the table illustrates, FusionLog’s F1‑score significantly outperforms all baselines. Specifically, PLELog, LogRobust, and NeuralLog are designed for single‑system anomaly detection; they lack the adaptability to bridge the gap between target and source systems. DeepLog trains using only a small number of normal logs from the target system, thus misclassifies novel normal events as anomalies. LogTransfer and LogTAD rely on simple transfer‑learning models that share only part of the neural network structure between source and target systems, and can achieve good performance only under specific conditions. LogDLR depends on adversarial domain adaptation and labeled logs from the target system, but when there is a substantial distributional shift between source and target domains, its generalization is severely limited. MetaLog is designed for target systems with few anomaly logs, but it is also constrained by the scarcity of labeled logs: if anomaly labels or all labels are removed, it cannot perform effective anomaly detection (for example, its F1‑score drops from 92.93 to 33.33). Unlike MetaLog, FreeLog, based on system‑agnostic representation meta‑learning, can detect anomalies without any target labeled logs; however, its performance is hampered by mismatches between general and proprietary knowledge, making it effective only on general logs. RAGLog, based on LLMs and retrieval‑augmented generation, is highly sensitive to prompt design and the quality of retrieved positive examples. For instance, without proper guidance, its F1‑score falls from 89.05 to 62.79. It is worth noting that RAGLog (g2) applies the LLM directly to the target system without any labeled logs or retrieval guidance, yet achieves only 45--49\% precision and 62--66\% F1-score. This result indicates that the LLM alone cannot reliably classify log anomalies from its pretrained knowledge, even on these widely studied public datasets. In contrast, FusionLog achieves over 94\% F1-score through the collaborative distillation mechanism, confirming that the performance gains originate from the method design rather than the LLM's prior exposure to these datasets. Overall, existing methods cannot solve the cold‑start problem. FusionLog, through training-free semantic routing and knowledge distillation and fusion based on LLM and small model, effectively fuses both general and proprietary knowledge, achieving efficient knowledge transfer from source system to target system.

\subsection{Ablation Studies}
\label{sec4.3}
We conducted ablation experiments on the HDFS to BGL setting to evaluate the contributions of each core module.

\begin{figure}[h!]
    \centering
    \includegraphics[width=1.0\columnwidth]{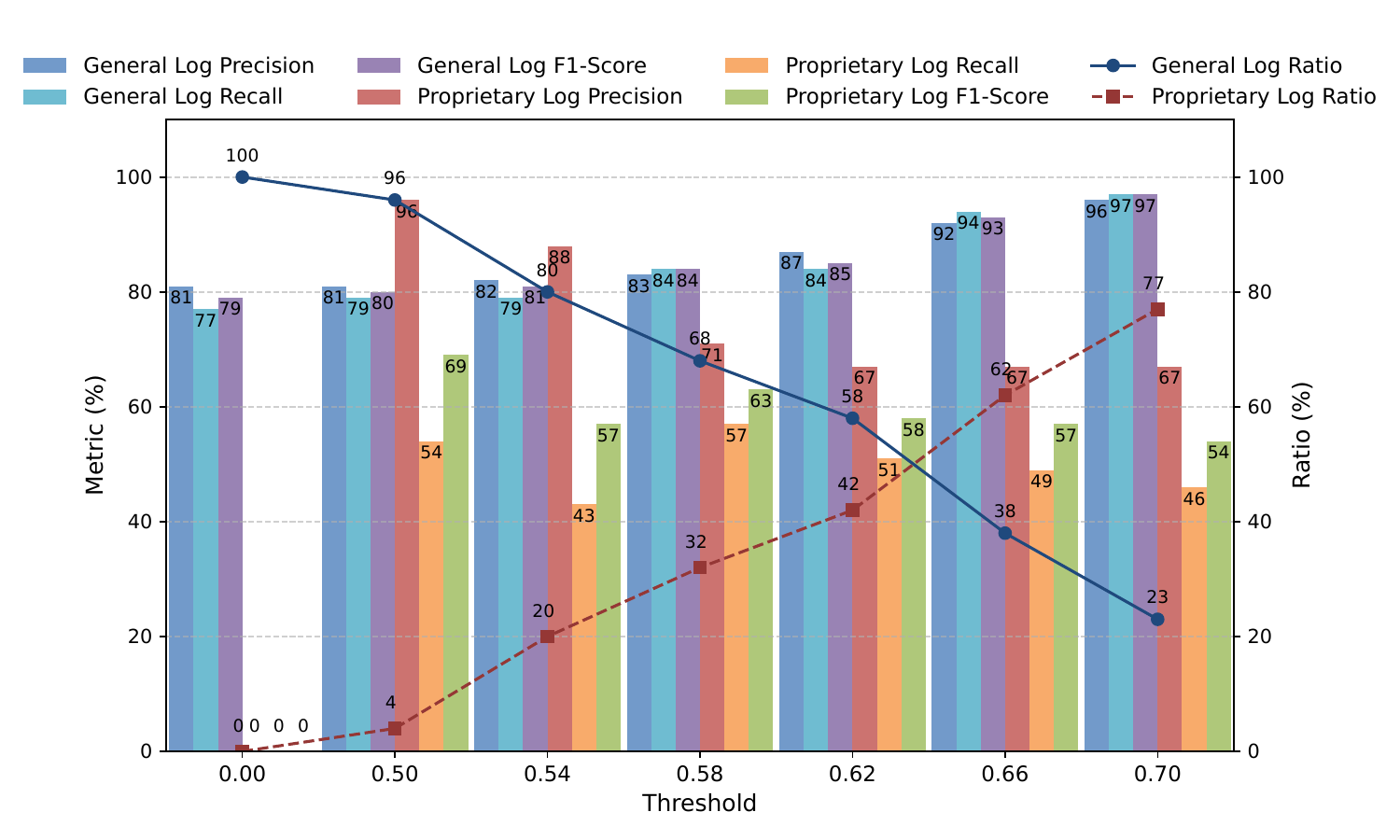}
    \caption{Routing Threshold Changes and Effects.}
    \label{fig_3}
\end{figure}

\textbf{Routing module.} To verify the necessity of the semantic routing module, we compare two configurations: (1) without routing, where the small model directly processes all target logs without distinguishing general from proprietary logs; and (2) with routing, where the router partitions target logs before differentiated processing. As shown in Figure \ref{fig_3}, without routing (threshold = 0), the small model achieves a moderate F1-score by treating all logs uniformly. After enabling routing, the small model maintains high accuracy on general logs but performs markedly worse on proprietary logs, confirming that a significant portion of target logs contains proprietary patterns that the small model alone cannot handle. This validates the necessity of the routing module: by explicitly separating general logs from proprietary logs, FusionLog can apply differentiated processing strategies, yielding substantial overall performance gains.

\begin{figure}[h!]
    \centering
    \includegraphics[width=1.0\columnwidth]{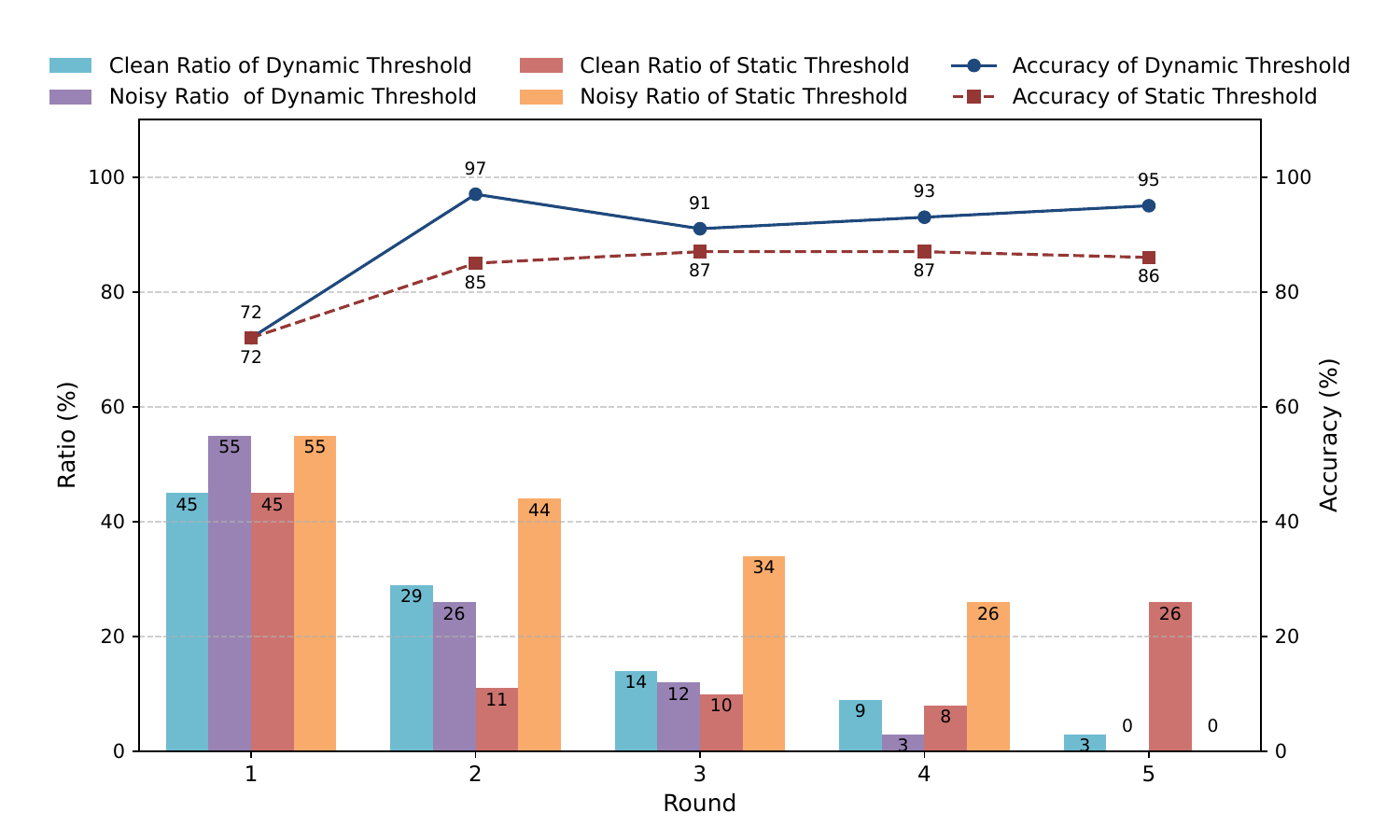}
    \caption{Round Variation and Threshold Strategy Comparison.}
    \label{fig_4}
\end{figure}

\textbf{Distillation module.} As shown in Figure \ref{fig_4}, the dynamic threshold strategy yields steady accuracy improvements across rounds as samples progressively migrate from the noisy pool to the clean pool, whereas a static threshold fails to deliver gains in later rounds, causing most samples to remain unprocessed until the final iteration. The small model's F1-score on the proprietary subset increases from 54 (before distillation) to 89 (after distillation), with Precision improving from 67 to 91 and Recall from 46 to 88, demonstrating effective proprietary knowledge injection. To assess error propagation, we measured the clean pool label precision at each round: Round 1 achieves 93.7\% and subsequent rounds consistently reach 96--99\%, confirming that the dual-verification mechanism effectively filters erroneous pseudo-labels. Injecting 10\% random noise into the clean pool decreases the final F1-score by only 1.8 percentage points (from 94.65 to 92.85), confirming robustness to moderate label noise through incremental dilution of early-round errors and the self-correcting RAG feedback loop.

\textbf{LLM robustness.} To evaluate the consistency of the collaborative distillation module across different LLMs, we replaced Qwen3 with DeepSeek-V3.2, GLM-4.6, and Gemini-2.5-Flash. The resulting F1-scores (95.71, 92.50, and 91.12, respectively) demonstrate that the module generalizes well across different LLMs under the support of RAG-based retrieval augmentation and few-shot prompting. The substantial performance variance across LLMs (ranging from 91.12 to 95.71) further suggests that FusionLog's effectiveness is driven by the collaborative distillation mechanism rather than any particular LLM's memorization of these public datasets, since all LLMs have comparable access to publicly available data yet yield markedly different results.

\textbf{Module-removal ablation.} To isolate the contribution of individual components, we conducted module-removal experiments. Removing the RAG retrieval module (i.e., using the LLM without in-context examples) reduces F1 from 94.65 to 87.91, confirming that retrieval-augmented prompting is essential for pseudo-label quality. Removing the dual-verification data selection (i.e., accepting all LLM pseudo-labels without small model filtering) degrades performance to 89.74, demonstrating that unfiltered noise accumulates across rounds. Removing the multi-round iterative mechanism (i.e., single-round distillation only) yields 90.83, highlighting the value of progressive refinement. All variants still outperform FreeLog (79.06), confirming that each component contributes positively while the full pipeline achieves the best performance through their synergistic combination.

\subsection{Parameter Sensitivity Analysis}
\label{sec4.4}

\textbf{Routing threshold $\tau$.} Having established the necessity of the routing module, we analyze the sensitivity of the similarity threshold $\tau$ in the HDFS to BGL setting. As shown in Figure \ref{fig_3}, as the threshold increases, the proportion of general logs decreases while proprietary logs increase, and the F1-score on general logs improves markedly, indicating that stricter criteria yield purer general logs. However, threshold selection involves a trade-off: a lower threshold may include proprietary patterns in the general branch and cause negative transfer, while a higher threshold over-assigns samples to the proprietary branch and increases the distillation burden.

\textbf{Routing quality validation.} Although no ground-truth labeling of general versus proprietary logs exists, routing quality can be indirectly assessed through downstream performance. At the default threshold, the small model achieves F1-score above 93\% on general logs but only 54\% on proprietary logs. This gap of nearly 40 percentage points confirms that the router effectively separates source-aligned patterns from system-specific ones: if proprietary logs were frequently misclassified as general, the contamination would substantially degrade the small model's performance on the general logs.

\textbf{Dynamic confidence threshold $\epsilon$.} We varied the initial threshold $\epsilon_0 \in \{0.80, 0.85, 0.90, 0.95\}$ and per-round step size $\Delta\epsilon \in \{0.02, 0.05, 0.10\}$. The F1-score remains stable within 92.53--94.65\% across all twelve configurations, with the default ($\epsilon_0=0.9$, $\Delta\epsilon=0.05$) achieving the best performance. Setting $\epsilon_0$ too low degrades performance due to early inclusion of unreliable samples, while setting it too high delays clean sample accumulation. Similarly, an overly large $\Delta\epsilon$ causes premature relaxation, whereas a small $\Delta\epsilon$ leaves too many samples unprocessed. The maximum variance across all configurations is within approximately 2 percentage points, confirming that the method is not highly sensitive to these parameters within reasonable ranges.

\subsection{Run Time Analysis}
\label{sec4.5}

\textbf{Training and inference time.} We conducted timing experiments on a single NVIDIA 3090 GPU in the HDFS to BGL setting. Among small model baselines, training times range from 7.5 minutes (LogRobust) to 19 minutes (LogTAD), with MetaLog and LogTransfer both requiring around 13 minutes, and PLELog around 9.5 minutes. All small model methods achieve inference times of approximately 5 seconds. FusionLog requires approximately 22 minutes for training due to the multi-round collaborative distillation and semantic routing, while its inference time (5 seconds, or 0.084 ms per sequence) remains comparable to all small model methods. In contrast, RAGLog requires no training but incurs a testing time of about 33 minutes due to the retrieval process and LLM inference overhead during deployment.

\textbf{Computational complexity.} The time complexity of the similarity computation in semantic routing is $\mathcal{O}(|x_k|\cdot |V^{\text{source}}|)$, where $|x_k|$ denotes the number of events in a log sequence and $|V^{\text{source}}|$ represents the number of events in the source system, whose practical scale typically falls within $10^1\sim10^2$, making this procedure computationally efficient. Moreover, this computation is only involved in the training stage for log partitioning and knowledge fusion. During inference, FusionLog directly performs end-to-end prediction using the fused small model without introducing any additional overhead from semantic routing or LLM invocation.

Although FusionLog incurs a moderate increase in training cost compared to small model methods, its testing time and inference overhead in practical deployment remain significantly lower than LLM-based approaches (e.g., RAGLog) and are comparable to other small model methods. More importantly, FusionLog operates in a true cold-start scenario without target-system labels, thereby substantially reducing annotation costs.

\section{Related Work}
Anomaly detection in software systems has garnered widespread attention, and various methods have been proposed. In single-system scenarios, DeepLog~\cite{10.1145/3133956.3134015} employs LSTM networks to model normal log event index sequences and identifies anomalies when new logs deviate from these learned patterns. LogAnomaly~\cite{ijcai2019p658}, on the other hand, enhances anomaly detection by leveraging Word2Vec embedding techniques to extract sequence and quantitative features from log events. LogRobust~\cite{10.1145/3338906.3338931} employs TF-IDF and word vectorization techniques to convert logs into semantic vectors. In this way, updated logs can be transformed into semantic vectors and incorporated into the model’s training and inference process. PLELog~\cite{9401970} introduces a semi-supervised learning model that derives log labels through unsupervised clustering methods, which are then used to build a supervised model. NeuralLog~\cite{9678773} avoids errors and information loss that may occur in traditional log parsing by directly processing raw logs. This method uses deep learning techniques to understand and analyze patterns and anomalous behaviors in logs. RAGLog~\cite{10607047} leverages a Retrieval‑Augmented LLM integrated with a vector database in a QA‑style pipeline to detect anomalies. 

In cross-system scenarios, LogTransfer~\cite{9251092} proposes a transfer learning framework using GloVe representations and LSTM, sharing fully connected layers between source and target systems to enable cross-system knowledge transfer. LogTAD~\cite{10.1145/3459637.3482209} uses adversarial domain adaptation to align log distributions across systems, thereby reducing reliance on target logs. LogFormer~\cite{10.1609/aaai.v38i1.27764} constructs a Transformer-based pretraining and adapter fine-tuning pipeline with a Log-Attention module to improve cross domain generalization with fewer trainable parameters. MetaLog~\cite{10.1145/3597503.3639205} combines globally consistent semantic embeddings with meta-learning to generalize labeled knowledge from mature systems to new ones. CroSysLog~\cite{10992557} learns event-level neural representations on source systems and then adapts efficiently to target systems using only a small amount of labeled logs. LogMeta~\cite{SUN2026112781} integrates model-agnostic meta-learning with a hybrid language model to form a semi-supervised few-shot adaptation framework. FreeLog~\cite{FreeLog} unites unsupervised domain adaptation with meta-learning to learn system-agnostic representations that support zero-label generalization. LogDLR~\cite{10910212} combines universal sentence embeddings with a Transformer-based autoencoder and domain adversarial training to obtain domain-invariant latent representations. LogSynergy~\cite{LogSynergy} centers on LLM-based event interpretation and system unified feature extraction to standardize log syntax and separate system-specific and shared features. LogAction~\cite{11334497} combines transfer learning with active learning, using free energy and uncertainty-driven sampling to improve adaptation with minimal labeled data. Finally, LogMoE~\cite{11334514} presents a parse-free lightweight mixture-of-experts framework that trains multiple experts on multi-source labeled logs and integrates them via a gating mechanism to achieve scalable and robust generalization to target systems.

However, existing studies concentrate on extracting general knowledge and overlook the significant discrepancies at the level of proprietary knowledge between the source and target systems, which motivates the research in this paper.

\section{Discussion and Limitations}

As demonstrated in our experiments, properly distinguishing and integrating general and proprietary knowledge is crucial for cross-system anomaly detection under zero-label conditions. By consolidating both types of knowledge into a small model, FusionLog maintains high detection accuracy while ensuring computational efficiency, making it practical for both production environments and trial-run phases.

We acknowledge that leveraging LLMs to generate pseudo-labels inevitably introduces a certain degree of noise, which may propagate through the iterative distillation process and affect model performance. To mitigate this risk, FusionLog incorporates several mechanisms: (1) the dual-verification data selection module filters samples by requiring agreement between LLM and small model predictions along with a confidence threshold, effectively reducing the proportion of erroneous pseudo-labels entering the training set; (2) the dynamic threshold strategy progressively lowers the confidence bar as the small model improves, preventing premature inclusion of unreliable samples in early rounds while avoiding training stagnation in later rounds; (3) the iterative enrichment of the RAG knowledge base with high-quality clean samples continuously improves the LLM's contextual references, thereby enhancing pseudo-label quality in subsequent rounds. Although these mechanisms cannot entirely eliminate pseudo-label noise, our experimental results demonstrate that the residual errors do not significantly degrade overall detection performance. Additionally, the current semantic routing relies on a single similarity threshold, which may not optimally partition logs when the semantic boundary between general and proprietary events is ambiguous.

\section{Conclusion}
In this paper, we propose FusionLog, a novel cross‐system log-based anomaly detection method designed to overcome performance limitations arising from the discrepancy between general knowledge and proprietary knowledge. Specifically, we introduce a training-free semantic router and a multi‐round collaborative knowledge distillation and fusion method based on LLM and small model to effectively fuse these two types of knowledge and achieve zero‐label generalization. Under a fully zero‐label setup, FusionLog attains an F1‐score exceeding 90\%, significantly outperforming state‐of‐the‐art cross‐system methods.

Building on these promising results, we outline several directions for future research. First, we aim to develop adaptive or learnable routing strategies to replace the current single-threshold mechanism, enabling more precise partitioning when the semantic boundary between general and proprietary events is ambiguous. Second, we plan to incorporate proprietary knowledge from multiple modalities, including source code, documentation, and configuration files, to better capture system-specific anomalies. Third, extending FusionLog to multi-source transfer scenarios, where labeled logs from multiple mature systems are jointly leveraged, is another promising direction.




\bibliographystyle{IEEEtran}
\bibliography{reference}


 




\vfill

\end{document}